\documentclass[runningheads]{llncs}
\usepackage[T1]{fontenc}
\usepackage{amsmath}
\usepackage{algorithmic}
\usepackage{multirow}
\usepackage{booktabs} 
\usepackage{subcaption}
\usepackage{amssymb} 
\usepackage{enumitem}
\usepackage{arydshln}
\usepackage[table]{xcolor}

\usepackage{graphicx}
\begin{document}
\title{SPOC-SQL: Stage-wise Preference Optimization for
Controllable Text-to-SQL\vspace*{-15pt}}
\titlerunning{SPOC-SQL}

\author{
Yingnan Chen\textsuperscript{\ensuremath{\dagger}} \and
Chun Ding\textsuperscript{\ensuremath{\dagger}} \and
Tianshi Xu \and
Xu Yang\textsuperscript{*} \and
Si Wu
}

\authorrunning{Y. Chen et al.}

\institute{
South China University of Technology,
Guangzhou, Guangdong 510006, P.R. China\\
\email{
chen.m0rem@gmail.com,
csdingchun@mail.scut.edu.cn,
\{xtshi,yxu8\}@grgbanking.com,
cswusi@scut.edu.cn
}\\[-1pt]
\textsuperscript{\ensuremath{\dagger}}
Joint first authors.
\quad
\textsuperscript{*}
Corresponding author.
}

\maketitle


%
\vspace*{-16pt}
\begin{abstract}
Text-to-SQL aims to translate natural language questions into executable SQL queries over relational databases, requiring multi-stage structured reasoning over database schemas and query constraints.
However, existing methods treat this task as single-step generation, where models optimize entire SQL sequences without targeted feedback at key decision points and lack support for interacting with and controlling the intermediate generation process.
To address this issue, we propose SPOC-SQL, which decomposes Text-to-SQL into four sequential subtasks following standard SQL execution logic and designs stage-specific optimization strategies for the model to learn key decisions. Specifically, we propose the implementation of fine-grained preference optimisation at key decision points across SQL stages, with the objective of enhancing structured decision-making during query construction.  Furthermore, a structured decomposition strategy is designed, facilitating stage-wise intervention and correction through explicit intermediate representations. This results in more controllable and reliable SQL generation.
Experiments demonstrate that incorporating stage-wise human knowledge consistently improves performance, validating the effectiveness of stage perception controllable generation.


%
\vspace*{-5pt}
\end{abstract}

\vspace*{-25pt}
\section{Introduction}
\vspace*{-7pt}

Text-to-SQL (T2S) is the process of translating natural language questions into SQL queries that function on relational databases. T2S tasks frequently require handling multi-table joins, nested subqueries, and complex conditions \cite{xie2026fcidiff}, thereby forming a multi-step decision-making process similar to multi-turn QA scenarios \cite{ip-kgqa}. Modeling this process accurately is challenging due to the complex relationships in the database schema and the semantic complexities involved.

Existing methods generate SQL sequences by jointly encoding questions and database schemas. Examples include RAT-SQL\cite{RAT-SQL}, which employs relation-aware attention, BRIDGE\cite{Bridge}, which utilizes tagged sequence encoding, and RESDSQL\cite{RESDSQL}, which applies decoupled schema linking.
However, these methods typically treat the entire SQL sequence as the optimization objective and learn it holistically without stage-level modeling, due to the scarcity of high-quality multi-turn T2S data. 
Specifically, when the model assigns equal importance to both key interaction steps that drive task progress and irrelevant conversational content, the signals required for effective decision-making are diluted. This approach hinders the model's ability to master core decisions that advance the task and restricts its capacity to develop essential interaction strategies \cite{cai2026text}.
Furthermore, the absence of structured modeling impedes interpretability, making it harder to trace errors to specific reasoning stages. Limited interpretability also constrains user engagement with intermediate query components, resulting in a reliance on either complete acceptance of the generated SQL or only coarse-grained modifications.

\begin{figure*}[t]
    \centering  
    \includegraphics[width=0.95\textwidth]{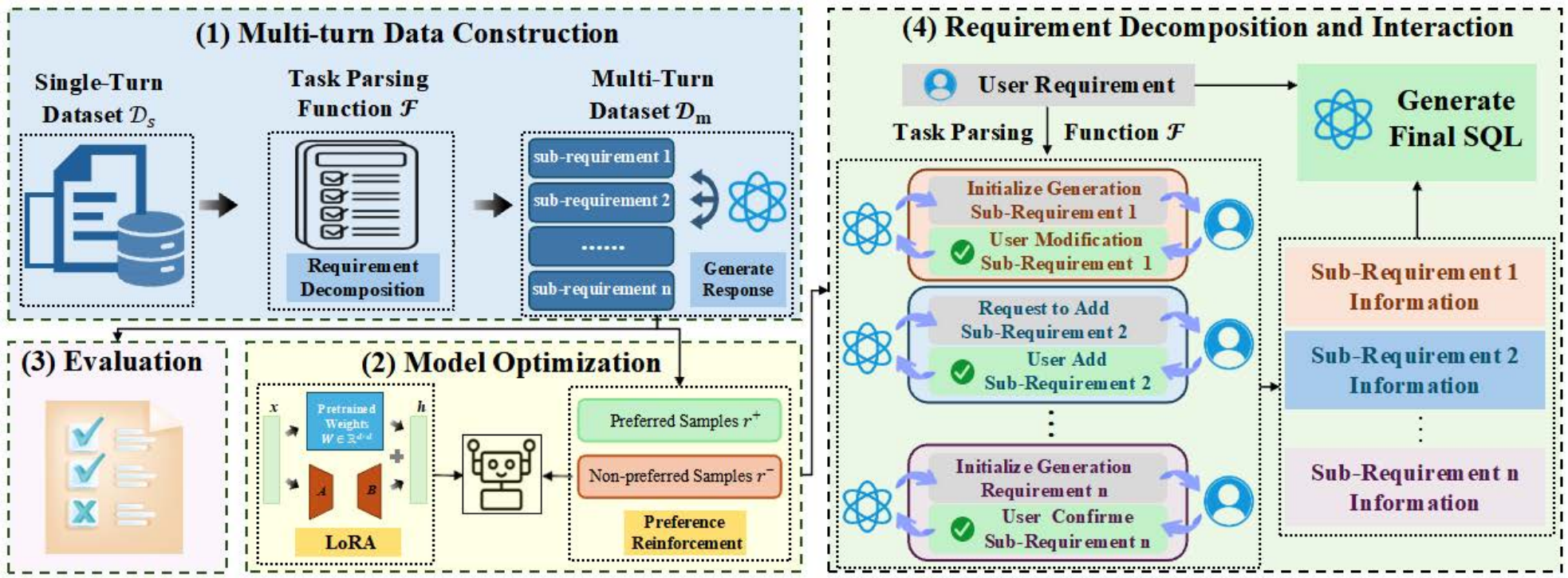}
    \caption{The workflow of SPOC-SQL. 
(1) The task parsing function $\mathcal{F}$ decomposes single-turn queries into multi-turn sequences with stage-specific sub-requirements. 
(2) LoRA fine-tuning combined with preference-based optimization to enhance decision capability through fine-grained policy optimization. 
(3) Evaluation of multi-turn QA capabilities. 
(4) At inference, $\mathcal{F}$ decomposes user requirements into sub-tasks for iterative confirmation, generating the final SQL.}
    \label{fig:overview}
    \vspace*{-16pt}
\end{figure*}


To address this issue, we propose SPOC-SQL, which decomposes T2S into four sequential subtasks aligned with standard SQL execution logic and employs stage-wise optimization strategies to enable the model to learn key decisions.
Specifically, SPOC-SQL decomposes T2S into structured subtasks corresponding to SQL execution stages, enabling progressive decision-making rather than single-step generation.
Single-turn instances are decomposed into sequential sub-stages to construct multi-turn QA samples. Based on this dataset, we propose Preference-based Multi-turn QA Decision Optimization (PMDO), which builds preference pairs at key decision points across subtasks. Preferred samples correspond to correct intermediate decisions derived from the gold SQL execution process, while non-preferred samples are automatically generated through prompt-guided controlled perturbations using the LLM \cite{ding2026llm}. Specifically, the model is prompted to intentionally modify critical intermediate reasoning elements, including selecting irrelevant columns, introducing incorrect filtering conditions \cite{zhang2025class}, mismatching aggregation operations, and applying improper sorting constraints, while preserving the overall query context and linguistic plausibility.
Additionally, PMDO integrates LoRA \cite{hu2022lora} with DPO \cite{dpo}, enabling efficient stage-level preference learning and improving decision discrimination.
We further design a Requirement Decomposition and Interaction Module (RDIM), which decomposes queries into stage-specific subtasks. Intermediate results are exposed for verification or correction, and the final SQL is synthesized from validated outputs, thereby improving interpretability and controllability.
The primary contributions of our work are summarized as follows:
\begin{itemize}[label=\textbullet]
    \vspace*{-3pt}
    \item Different from existing methods that approach Text-to-SQL as a single-step sequence generation task, we propose SPOC-SQL, a framework that models the task as a stage-wise structured decision process. Stage perception strategies are integrated into both training and inference, which facilitates unified fine-grained preference optimization and controllable generation.
    \item To enhance decision accuracy on core subtasks, we design a stage-level preference optimization strategy, PMDO, which extends DPO supervision from the sequence level to key decision points across distinct SQL logical stages.
    \item We propose RDIM to enable stage-wise verification and correction during inference, endowing the SQL generation process with explicit interpretability and user-driven intervention capabilities.
\end{itemize}

\vspace*{-10pt}
\section{Methodology}
\vspace*{-8pt}
\subsection{Overview} 
\vspace*{-5pt}
The workflow of SPOC-SQL is outlined in Fig.~\ref{fig:overview}. Multi-turn Data Construction uses a task-parsing function $\mathcal{F}(\cdot)$ to decompose natural language queries into stage-specific subtasks, such as table and column selection, condition filtering, group aggregation, and ordering or limiting. This approach transforms single-turn queries into multi-turn interaction sequences that capture incremental user intent expression, supplementation, and correction.
Model Optimization, based on the constructed dataset $\mathcal{D}_{\text{m}}$, incorporates preference constraints at key decision-making nodes in each interaction turn. The model is updated through a policy optimization paradigm that integrates LoRA-based efficient fine-tuning with DPO. This process enhances decision-making capability at key interactive steps and improves SQL generation accuracy in multi-turn scenarios.
Evaluation assesses model performance across multi-turn dialogue completion, single-turn reasoning, and error correction, thereby capturing both global consistency and local decision accuracy.
Finally, Requirement Decomposition and Interaction is designed for inference, where user queries are decomposed into structured sub-tasks and intermediate results are iteratively refined through user feedback, enabling a controllable and correctable SQL generation process for complex queries.

\vspace*{-8pt}
\subsection{Multi-turn Data Construction and Evaluation Module}
\vspace*{-4pt}
\subsubsection{Multi-turn data construction.}  
T2S queries expressed in natural language often correspond to multiple stage-specific sub-tasks aligned with SQL execution logic. To explicitly model stage-wise structure, we define a task parsing function $\mathcal{F}(\cdot)$ that decomposes a QA sample into several predefined stages following the canonical SQL execution pipeline, namely SELECT-FROM, WHERE, GROUP-HAVING, and ORDER-LIMIT.
Given a single-turn QA sample $(x, y)$, where $x$ denotes a natural language query and $y$ represents the corresponding SQL statement, the decomposition is expressed as:
\begin{equation}
(s^{\text{SF}}, s^{\text{WH}}, s^{\text{GH}}, s^{\text{OL}}) = \mathcal{F}(x),
\end{equation}
where $s^{\text{SF}}$ represents table and column selection (SELECT-FROM), $s^{\text{WH}}$ represents condition filtering (WHERE), $s^{\text{GH}}$ represents grouping and aggregation (GROUP-HAVING), and $s^{\text{OL}}$ represents ordering and limits (ORDER-LIMIT). Sub-tasks absent in a query may be left empty or skipped during dataset construction.  
The single-turn query is transformed into a multi-turn interaction sequence. At turn $t$, user input focuses on the current stage sub-task while maintaining overall semantic consistency with the complete query and standard SQL. The model generates the intermediate result conditioned on previous context:
\begin{equation}
\begin{aligned}
u^{(t)} &= (s^{(t)}, x, y), \\
r^{(t)} &= \mathcal{M}(u^{(t)}, \mathbf{h}^{(t-1)}),
\end{aligned}
\end{equation}
where $\mathcal{M}(\cdot)$ denotes the large language model (LLM) that produces $r^{(t)}$ conditioned on the previous context $\mathbf{h}^{(t-1)}$. The final turn $T$ directly adopts the original standard SQL as the output, $r^{(T)} \equiv y$, ensuring consistency between multi-turn generated results and the original annotations.  
Different query types are simulated to reflect realistic user behavior. Explicit-demand queries provide complete conditions in the initial turn, resulting in fewer dialogue turns. Ambiguous-demand queries involve stage-wise supplementation of filter conditions or aggregation requirements, forming an iterative refinement process. Error-correction queries introduce inconsistent or erroneous conditions in some turns, corrected in subsequent turns to enhance the model's capability for intention correction.  
The multi-turn dataset is formally defined as:
\begin{equation}
\mathcal{D}_{\text{m}} = \{\mathcal{C}_{i}\}_{i=1}^{N}, \quad
\mathcal{C}_i = \{(u_i^{(t)}, y_i^{(t)})\}_{t=1}^{T_i},
\end{equation}
where each $\mathcal{C}_{i}$ represents a complete multi-turn query process. The conversion transforms static single-turn T2S data into dynamic multi-turn interaction sequences, preserving original semantics while explicitly modeling incremental user expression, supplementation, and correction. The constructed dataset contains three types of multi-turn interactions: explicit requests, where the full user requirement is provided in the first turn; vague requests, where the complete requirement is gradually revealed across multiple turns; and revised requests, where partial or incorrect sub-requirements are corrected in subsequent turns.

\vspace*{-8pt}
\subsubsection{Multi-turn QA evaluation.}  
The evaluation framework assesses model performance across three sub-tasks: multi-turn QA, single-turn QA, and error-correction tasks. Evaluation captures overall dialogue consistency, local context comprehension, and response to user correction instructions.  
The multi-turn QA sub-task requires completion of a full dialogue. At each turn $t$, the model generates a response based on current input and accumulated dialogue state:
\begin{equation}\label{eq:multi_turn_update}
\begin{aligned}
r_i^{(t)} &= \mathcal{M}(u_i^{(t)}, h_i^{(t-1)}), \\
h_i^{(t)} &= h_i^{(t-1)} \cup \{u_i^{(t)}, r_i^{(t)}\},
\end{aligned}
\end{equation}
where $t = 1, \dots, T_i$ denotes the turn index within the $i$-th dialogue consisting of $T_i$ turns, and $h_i^{(0)} = \varnothing$ indicates that the dialogue history is initially empty.  
$u_i^{(t)}$ and $r_i^{(t)}$ denote the user input and model response at turn $t$, respectively, and $h_i^{(t)}$ represents the accumulated dialogue history up to turn $t$.  
Evaluation relies on the final turn output $r_i^{(T_i)}$ compared to the standard SQL $y_i^{(T_i)}$, emphasizing global consistency in extended dialogues.  
The single-turn QA sub-task treats each turn independently to prevent error accumulation across turns. Historical context $h_i^{(t-1)}$ is provided directly from the dataset and remains unchanged. Evaluation across all turns measures the model's comprehension of local context and the execution of stage-specific sub-tasks.  
The error-correction QA sub-task focuses on turns containing explicit user corrections. Responses are generated based on corrected input $\tilde{u}_i^{(t)}$ and preceding context:
\begin{equation}
r_i^{(t)} = \mathcal{M}(\tilde{u}_i^{(t)}, h_i^{(t-1)}),
\end{equation}
where $t \in \mathcal{T}_i^{\text{err}}$ denotes the set of turns that involve explicit correction instructions from the user.
The evaluation setting measures the model's ability to identify and revise previously generated erroneous conditions based on user feedback.
The three-tier evaluation framework offers a comprehensive assessment of multi-turn T2S capability.

\vspace*{-8pt}
\subsection{Requirement Decomposition and Interaction Module}
\vspace*{-4pt}
\subsubsection{Requirement decomposition and extraction.}  
During inference, user queries expressed in natural language are mapped to SQL statements with explicit logical structure. The initial task information set is defined as:
\begin{equation}
\mathcal{I}_i^{(0)} = { (s_i^{\text{SF}}, r_i^{\text{SF}}), (s_i^{\text{WH}}, r_i^{\text{WH}}), (s_i^{\text{GH}}, r_i^{\text{GH}}), (s_i^{\text{OL}}, r_i^{\text{OL}}) },
\end{equation}
where stage-wise decomposition using $\mathcal{F}(\cdot)$ generates four sub-tasks, and the model produces the corresponding initial stage responses $r_i^{(k)}$. Stage responses enables stage-level interaction with the user for confirmation and correction, ensuring accurate and controllable task information.

\vspace*{-8pt}
\subsubsection{Interactive knowledge.}  
User confirmation or correction at each stage serves to refine the model's understanding of the task requirements. At turn $t$, the user reviews the intermediate outputs produced for each sub-task and provides feedback, which may include confirming correct predictions or correcting errors. The feedback updates the stage-specific sub-tasks $\tilde{s}_i^{(k)}$ and their corresponding responses $\tilde{r}_i^{(k)}$, resulting in the aggregated set of updated interactions:
\begin{equation}
	\mathcal{I}_i^{(t)} = \{ (\tilde{s}_i^{\text{SF}}, \tilde{r}_i^{\text{SF}}), (\tilde{s}_i^{\text{WH}}, \tilde{r}_i^{\text{WH}}), (\tilde{s}_i^{\text{GH}}, \tilde{r}_i^{\text{GH}}), (\tilde{s}_i^{\text{OL}}, \tilde{r}_i^{\text{OL}}) \},
\end{equation}
where each pair $(\tilde{s}_i^{(k)}, \tilde{r}_i^{(k)})$ captures the corrected sub-task state and the corresponding model output for the $k$-th stage, enabling a precise and user-aligned representation of the task at that point in the dialogue.  
The model subsequently leverages the complete task query $x_i$, together with the updated stage-specific information $\mathcal{I}_i^{(t)}$, to generate the final SQL statement:
\begin{equation}
	y_i = \mathcal{M}(x_i, \mathcal{I}_i^{(t)}),
\end{equation}
where the integration of $x_i$ with $\mathcal{I}_i^{(t)}$ allows the model to account for both the original user intent and the refinements provided through user interaction. The stage-wise guidance ensures that the generated SQL query faithfully reflects the user’s intentions while maintaining syntactic correctness, logical consistency, and adherence to the specific requirements of each sub-task. The interaction strategy establishes a controllable inference path, providing opportunities for user intervention and iterative correction throughout the generation process.

\vspace*{-8pt}
\subsection{Model Optimization}  
\vspace*{-4pt}
Following the construction of $\mathcal{D}_{\text{m}}$, the model undergoes task-specific optimization for stable execution of stage-wise decisions and generation of accurate SQL queries. The proposed PMDO approach combines LoRA for efficient fine-tuning with DPO.  
Each multi-turn sample $x_i \in \mathcal{D}_{\text{m}}$ is parsed into ordered stage sub-tasks $(s_i^{\text{SF}}, s_i^{\text{WH}}, s_i^{\text{GH}}, s_i^{\text{OL}})$, where each stage contains a set of key decision elements $\mathcal{U}_i^{(k)}$, including table and column selection, condition constraints,  aggregation/grouping, and ordering/limiting rules. The decision elements define the core decision space affecting semantic correctness of SQL generation.

LoRA fine-tuning updates linear layer weights $W_0$ with low-rank increments $\Delta W = \alpha BA$, freezing original parameters to preserve general model capabilities while adapting to stage-wise decision characteristics. Preference signals are constructed for each decision element $u \in \mathcal{U}_i^{(k)}$, forming pairs $(r_{i,u}^{+}, r_{i,u}^{-})$, where positive samples $r_{i,u}^{+}$ correspond to correct outputs and negative samples $r_{i,u}^{-}$ introduce controlled perturbations, such as incorrect tables, columns, conditions, aggregations, groupings, or ordering rules.  
Unlike standard preference optimization methods that operate at the sequence level, we perform preference learning at the stage-specific decision level. For each stage $k$, preference signals are conditioned on the corresponding sub-task context, enabling the model to distinguish fine-grained decision errors and learn stage perception decision boundaries. The optimization objective is formulated as:
\begin{equation}
\mathcal{L}_{\text{DPO}} = - \sum_{i,k} \sum_{u \in \mathcal{U}_i^{(k)}} \log \sigma \Bigg(
\log \frac{\pi_\theta(r_{i,u}^{+} \mid s_i^{(k)})}{\pi_{\text{ref}}(r_{i,u}^{+} \mid s_i^{(k)})}
-
\log \frac{\pi_\theta(r_{i,u}^{-} \mid s_i^{(k)})}{\pi_{\text{ref}}(r_{i,u}^{-} \mid s_i^{(k)})}
\Bigg),
\end{equation}
where $\pi_\theta$ denotes the policy model under optimization, $\pi_{\text{ref}}$ denotes the frozen reference model, and $s_i^{(k)}$ represents the context for stage $k$ of sample $i$, including both the user query and any outputs from previous SQL generation stages. Preference signals are introduced sequentially according to the SQL generation stage, with outputs from previous stages serving as context for subsequent stages. Stage-specific preference losses are accumulated to update LoRA parameters, establishing a generation bias aligned with task logic and ensuring coherence across multi-turn decision sequences.

\vspace*{-12pt}
\section{Experiments}
\vspace*{-7pt}
In this section, SPOC-SQL is evaluated on multiple Text-to-SQL benchmarks and a multi-turn dataset, analyzing its effectiveness in both single-turn and multi-turn scenarios. Dataset details, and implementation settings are presented, followed by comprehensive comparison with state-of-the-art methods, performance assessments across models and difficulty levels, ablation study, case study, user study, providing a thorough evaluation of the proposed method.
\vspace*{-13pt}
\subsection{Settings}
\vspace*{-6pt}
\subsubsection{Datasets.}
The Spider-Dev dataset \cite{spider} has 8,659 training instances across 200 databases. Spider-Realistic \cite{spider-realistic} removes explicit column references to test semantic understanding. In addition, we construct a multi-turn T2S dataset, T2S-MTD, consisting of 71,772 instances categorized into explicit, vague, and revised request interactions to simulate progressive user requirements in real business scenarios. Explicit requests provide sufficient information for direct SQL generation, vague requests require progressive supplementation across multiple turns, and revised requests involve modifying previous requirements during interaction. Evaluation uses execution accuracy (EX) and error correction (EC), measuring result equivalence and correction ability across turns.

\vspace*{-14pt}
\subsubsection{Implementation details.}
The LoRA uses a rank of $r=8$ and a scaling factor $\alpha=16$, and is injected into the self-attention layers and the linear projection layers of the feed-forward network. During fine-tuning, the original model is frozen, and only the LoRA parameters $A$ and $B$ are updated. The optimizer is AdamW with a learning rate of $\eta = 2\times10^{-4}$ and weight decay $w_{\text{decay}} = 0.01$. The single-device batch size is $B_{\text{single}}=2$, with gradient accumulation resulting in an effective batch size of $B_{\text{eff}}=32$. The maximum input length is $L_{\text{max}}=4096$, and training is conducted for $E=5$ epochs.
Competing methods include GPT-4 \cite{gpt4}, DeepSeek-V3 \cite{dpv3}, Qwen3 \cite{qwen3}, C3 \cite{C3}, ACT-SQL \cite{ACT-SQL}, DIN-SQL \cite{DIN-SQL}, DAIL-SQL \cite{dail}, MAC-SQL \cite{mac}, and MCS-SQL \cite{mcs}, with 4,096 context length.
We adopt a simulated human–computer interaction setting, where stage-wise intervention signals are constructed offline and progressively injected to simulate user corrections on intermediate representations, ensuring reproducibility.

\begin{table}[t]
\centering
\vspace*{-8pt}
\caption{Comparative performance on Spider benchmarks.}
\label{tab:spider-combined-results}

\begingroup
\setlength{\tabcolsep}{2.6pt}
\renewcommand{\arraystretch}{0.88}
\fontsize{6.2}{7.2}\selectfont

\begin{tabular}{llccccccc}
\toprule
\textbf{Method} &
\textbf{Model} &
\textbf{Venue} &
\textbf{SF} &
\textbf{WH} &
\textbf{GH} &
\textbf{OL} &
\textbf{Dev} &
\textbf{Real.} \\
\midrule

ACT-SQL~\cite{ACT-SQL}
& ChatGPT & EMNLP'23
& -- & -- & -- & --
& 80.4 & 75.8 \\

C3~\cite{C3}
& ChatGPT & arXiv'23
& -- & -- & -- & --
& 81.8 & 75.4 \\

DIN-SQL~\cite{DIN-SQL}
& GPT-4 & NeurIPS'23
& -- & -- & -- & --
& 82.8 & 78.1 \\

ACT-SQL~\cite{ACT-SQL}
& GPT-4 & EMNLP'23
& -- & -- & -- & --
& 82.9 & -- \\

DAIL-SQL~\cite{dail}
& DS-V3 & VLDB'24
& -- & -- & -- & --
& 83.2 & 77.2 \\

DAIL-SQL~\cite{dail}
& GPT-4 & VLDB'24
& -- & -- & -- & --
& 84.4 & 75.6 \\

MAC-SQL~\cite{mac}
& GPT-4 & COLING'25
& -- & -- & -- & --
& 86.8 & -- \\

MCS-SQL~\cite{mcs}
& GPT-4 & COLING'25
& -- & -- & -- & --
& 89.5 & -- \\

\midrule

\multirow{5}{*}{\textbf{SPOC-SQL}}
& \multirow{5}{*}{DS-V3}
& \multirow{5}{*}{--}
& -- & -- & -- & --
& 85.5 & 79.5 \\

& & &
$\checkmark$ & -- & -- & --
& 89.7 & 82.3 \\

& & &
$\checkmark$ & $\checkmark$ & -- & --
& 92.7 & 91.3 \\

& & &
$\checkmark$ & $\checkmark$ & $\checkmark$ & --
& 95.4 & 91.5 \\

& & &
\textbf{$\checkmark$} &
\textbf{$\checkmark$} &
\textbf{$\checkmark$} &
\textbf{$\checkmark$}
& \textbf{95.6} &
\textbf{93.1} \\

\bottomrule
\end{tabular}

\endgroup
\vspace*{-8pt}
\end{table}

\vspace*{-10pt}
\subsection{Comparison with State-of-the-Arts}
\vspace*{-6pt}
We evaluate the performance of the proposed SPOC-SQL method on the Spider-Dev and Spider-Realistic benchmark datasets against competing Text-to-SQL methods.
In Table \ref{tab:spider-combined-results}, ``$-$'' in the columns SF, WH, GH, or OL indicates that human knowledge was not incorporated at the corresponding stage, while ``$\checkmark$'' indicates that human knowledge was progressively introduced at that stage.
With incremental intervention of human knowledge, performance improves consistently and substantially. Introducing knowledge at the schema-focused stage (SF) raises EX to 89.7\%, exceeding MCS-SQL (89.5\%), a Competing method. Further incorporation of additional components leads to continued gains, ultimately reaching 95.6\% on Spider-Dev and 93.1\% on Spider-Realistic with full-component integration.
The results demonstrate that SPOC-SQL provides a competitive performance without external knowledge and achieves significant improvements through human knowledge intervention. The proposed method decomposes complex Text-to-SQL generation tasks into multiple structured sub-tasks, reducing overall task complexity while enabling precise human knowledge intervention at different stages. Such a design not only improves controllability and interpretability but also allows targeted correction of intermediate errors. Consistent and substantial performance gains across both benchmark datasets clearly highlight the effectiveness, scalability, and overall robustness of the proposed method in improving Text-to-SQL generation performance.

\begin{table}[t]
\centering
\caption{Performance by difficulty on Spider-Dev and Spider-Realistic.}
\label{tab:spider-combined-difficulty}

\begingroup
\setlength{\tabcolsep}{1.9pt}
\renewcommand{\arraystretch}{0.86}
\fontsize{6.0}{7.0}\selectfont

\begin{tabular}{lccccccccccc}
\toprule
\textbf{Method} &
\textbf{Data} &
\textbf{Model} &
\textbf{SF} &
\textbf{WH} &
\textbf{GH} &
\textbf{OL} &
\textbf{Easy} &
\textbf{Med.} &
\textbf{Hard} &
\textbf{X-Hard} &
\textbf{All} \\
\midrule

MCS-SQL~\cite{mcs}
& Dev & GPT-4
& -- & -- & -- & --
& 94.0 & 93.5 & 88.5 & 72.9 & 89.5 \\

DAIL-SQL~\cite{dail}
& Dev & DS-V3
& -- & -- & -- & --
& 93.5 & 83.0 & 83.9 & 67.5 & 83.2 \\

\cdashline{1-12}[1pt/1pt]

\multirow{5}{*}{\textbf{SPOC-SQL}}
& \multirow{5}{*}{Dev}
& \multirow{5}{*}{DS-V3}
& -- & -- & -- & --
& 95.2 & 84.3 & 85.6 & 74.1 & 85.5 \\

& & &
$\checkmark$ & -- & -- & --
& 96.0 & 90.1 & 89.1 & 79.5 & 89.7 \\

& & &
$\checkmark$ & $\checkmark$ & -- & --
& 97.2 & 91.9 & \textbf{97.1} & 83.7 & 92.7 \\

& & &
$\checkmark$ & $\checkmark$ & $\checkmark$ & --
& 97.6 & 96.6 & 96.6 & 87.5 & 95.4 \\

& & &
\textbf{$\checkmark$} &
\textbf{$\checkmark$} &
\textbf{$\checkmark$} &
\textbf{$\checkmark$}
& \textbf{97.6} &
\textbf{96.6} &
96.6 &
\textbf{88.6} &
\textbf{95.6} \\

\midrule

DAIL-SQL~\cite{dail}
& Real. & DS-V3
& -- & -- & -- & --
& 87.2 & 81.8 & 74.7 & 58.8 & 77.2 \\

\cdashline{1-12}[1pt/1pt]

\multirow{5}{*}{\textbf{SPOC-SQL}}
& \multirow{5}{*}{Real.}
& \multirow{5}{*}{DS-V3}
& -- & -- & -- & --
& 92.7 & 82.8 & 74.7 & 62.9 & 79.5 \\

& & &
$\checkmark$ & -- & -- & --
& 89.0 & 86.2 & 79.8 & 69.1 & 82.3 \\

& & &
$\checkmark$ & $\checkmark$ & -- & --
& \textbf{98.2} & 93.1 & 91.9 & 79.4 & 91.3 \\

& & &
$\checkmark$ & $\checkmark$ & $\checkmark$ & --
& 96.8 & 95.5 & 91.9 & 81.5 & 92.4 \\

& & &
\textbf{$\checkmark$} &
\textbf{$\checkmark$} &
\textbf{$\checkmark$} &
\textbf{$\checkmark$}
& 97.2 &
\textbf{96.0} &
\textbf{92.1} &
\textbf{83.5} &
\textbf{93.1} \\

\bottomrule
\end{tabular}

\endgroup
\vspace*{-10pt}
\end{table}

Spider-Dev and Spider-Realistic are further divided into four difficulty levels, namely Easy, Medium, Hard, and Extra Hard, to provide a more fine-grained evaluation of model performance. Table \ref{tab:spider-combined-difficulty} presents the detailed results under different difficulty settings.
On the Extra Hard subset of Spider-Dev, SPOC-SQL achieves 74.1\% EX even without human knowledge intervention, surpassing the current best method, MCS-SQL (72.9\%). Such improvement can be attributed to the decomposition of complex Text-to-SQL tasks into multiple structured sub-tasks, which reduces task complexity and enables more accurate intermediate reasoning.
With human knowledge intervention, performance improves across all difficulty levels. On Spider-Realistic, performance on the Extra Hard subset increases from 62.9\% to 83.5\%, demonstrating consistent gains under more challenging and realistic conditions \cite{zhang2026classbooth}.
Results show that the proposed requirement decomposition and interaction module effectively simplify complex queries and support targeted human intervention, leading to consistent performance improvements on complex Text-to-SQL tasks.


\vspace*{-10pt}
\subsection{Performance of SPOC-SQL Across Models and Datasets}
\vspace*{-6pt}

\begin{figure}[t]
    \centering  
    \includegraphics[width=0.95\columnwidth]{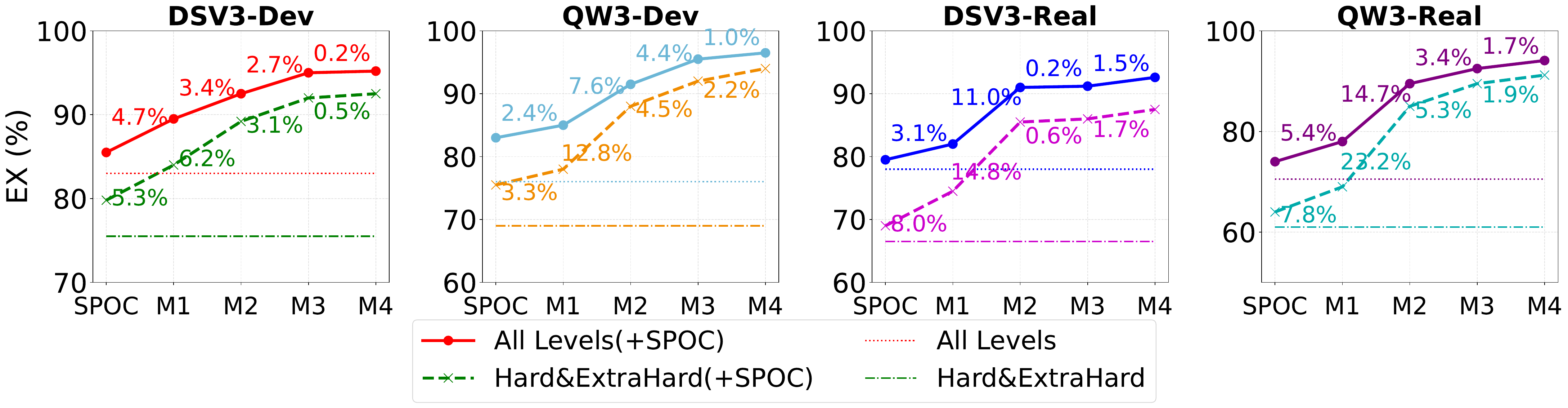}
    \vspace*{-10pt}
    \caption{Performance of SPOC-SQL across different datasets and models: Evaluations on DeepSeekV3(DSV3) and Qwen3-72B(QW3) across Spider-Dev(Dev) and Spider-Realistic(Real) Datasets.}
    \label{fig:gt}
    \vspace*{-13pt}
\end{figure}
To evaluate the performance of SPOC-SQL with different large language models on different datasets, we present an experiment in Fig. \ref{fig:gt}, where the method is tested with DeepSeek-V3 and Qwen3-72B models on Spider-Dev and Spider-Realistic datasets. We compare the base SPOC (without human intervention) and four progressively enhanced configurations: M1 with human intervention in the SF stage, M2 adding WH stage, M3 further incorporating GH stage, and M4 with intervention in all four stages (SF, WH, GH, and OL). Performance improvements are consistently observed across all model-dataset combinations when incorporating interactive components, with the magnitude of improvement on the Hard \& Extra Hard subsets exceeding that on all levels. The results indicate that SPOC-SQL effectively improved performance on complex queries and demonstrate the method's generalizability across different models and datasets.

\begin{figure}[t]
\centering
\includegraphics[width=0.95\linewidth]{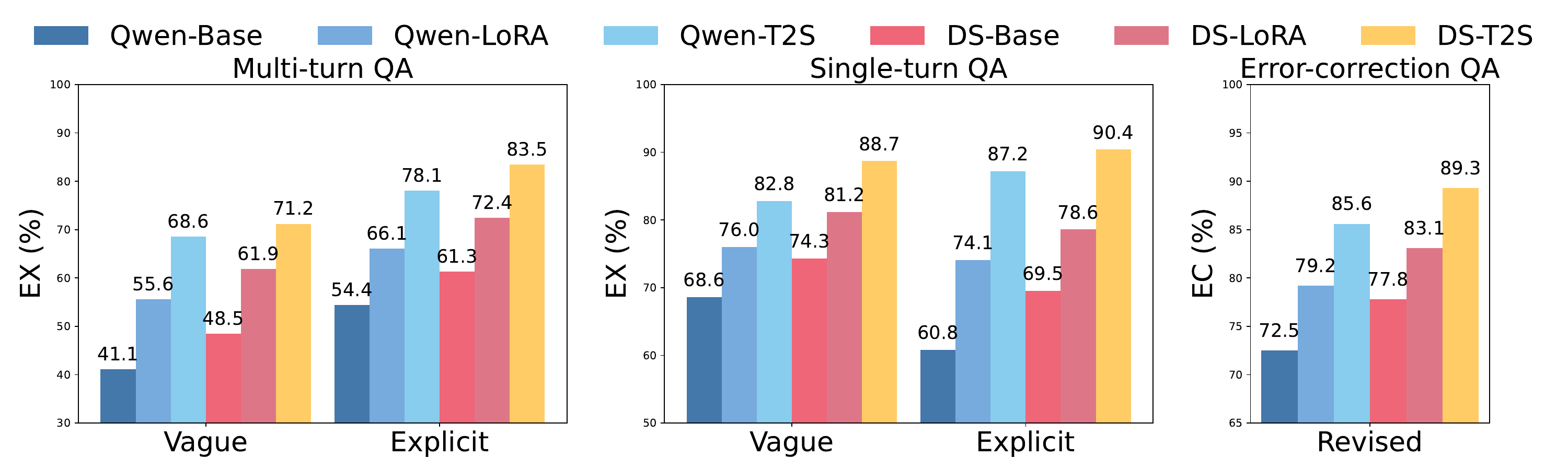}
\vspace*{-12pt}
\caption{\label{fig:3-6multi_model}Comparison of models on multi-turn QA, single-turn QA, and error-correction QA tasks.}
    \vspace*{-21pt}
\end{figure}

We further evaluate SPOC-SQL on the multi-turn T2S dataset T2S-MTD using Qwen3-72B and DeepSeek-V3. Base models without fine-tuning are Qwen-Base and DS-Base. Fine-tuning is done either with LoRA on single-turn Spider dataset (-LoRA) or with LoRA combined with the proposed optimization method on T2S-MTD (-T2S), enabling fair comparison and isolating the effect of multi-turn reinforcement. Evaluation follows the multi-turn QA, single-turn QA, and error-correction QA tasks designed in the prior work.
As shown in Fig.~\ref{fig:3-6multi_model}, Qwen-T2S outperforms Qwen-Base and Qwen-LoRA on multi-turn QA, single-turn QA, and error-correction QA tasks, while DS-T2S further surpasses Qwen-T2S under the same training strategy. The results confirm that SPOC-SQL’s task-decomposition and stage-wise optimization consistently improve performance and generalize across models.


\begin{table}[t]
\centering
\vspace*{-10pt}
\caption{Ablation study of SPOC-SQL.}
\label{tab:ablation}

\begingroup
\setlength{\tabcolsep}{5pt}
\renewcommand{\arraystretch}{0.85}
\fontsize{6}{7}\selectfont

\begin{tabular}{lccc}
\toprule
\textbf{Model} &
\textbf{Spider-Dev} &
\textbf{Spider-Real.} &
\textbf{T2S-MTD} \\
\midrule
\textbf{SPOC-SQL} & \textbf{95.6} & \textbf{93.1} & \textbf{84.6} \\
\midrule
w/o PMDO & 93.2 & 90.1 & 80.7 \\
w/o RDIM & 87.5 & 84.6 & 78.4 \\
LoRA-only & 83.7 & 79.8 & 75.4 \\
Base & 80.1 & 74.5 & 66.3 \\
\bottomrule
\end{tabular}

\endgroup
\vspace*{-8pt}
\end{table}

\vspace*{-14pt}
\subsection{Ablation Study}
\vspace*{-6pt}
To evaluate the contributions of SPOC-SQL, we conduct an ablation study based on DeepSeek-V3 on Spider-Dev, Spider-Realistic, and the proposed multi-turn T2S-MTD dataset. The evaluated modules include: (1) Preference-driven Multi-turn QA Decision Optimization (PMDO), which introduces stage-wise preference optimization during training; and (2) Requirement Decomposition and Interaction Module (RDIM), which decomposes SQL generation into stage-wise sub-tasks during inference.
We consider the following variants: \textbf{SPOC-SQL} employs both PMDO and RDIM; \textbf{w/o PMDO} denotes SPOC-SQL without PMDO while keeping RDIM; \textbf{w/o RDIM} denotes SPOC-SQL without RDIM while keeping PMDO; \textbf{LoRA-only} applies LoRA on the single-turn dataset Spider without proposed modules; and \textbf{Base} represents the base model without fine-tuning or proposed modules.
As shown in Table~\ref{tab:ablation}, removing either PMDO or RDIM degrades performance on both Spider benchmarks and T2S-MTD. Full SPOC-SQL achieves the best overall results, while LoRA-only improves over the base model but remains inferior to SPOC-SQL. These results demonstrate the effectiveness of stage-wise decomposition and preference-driven optimization.

\vspace*{-14pt}
\subsection{Case Study}
\vspace*{-6pt}
\begin{figure}[t]
    \centering
    \begin{subfigure}[t]{0.495\columnwidth}
        \centering
        \includegraphics[width=\linewidth]{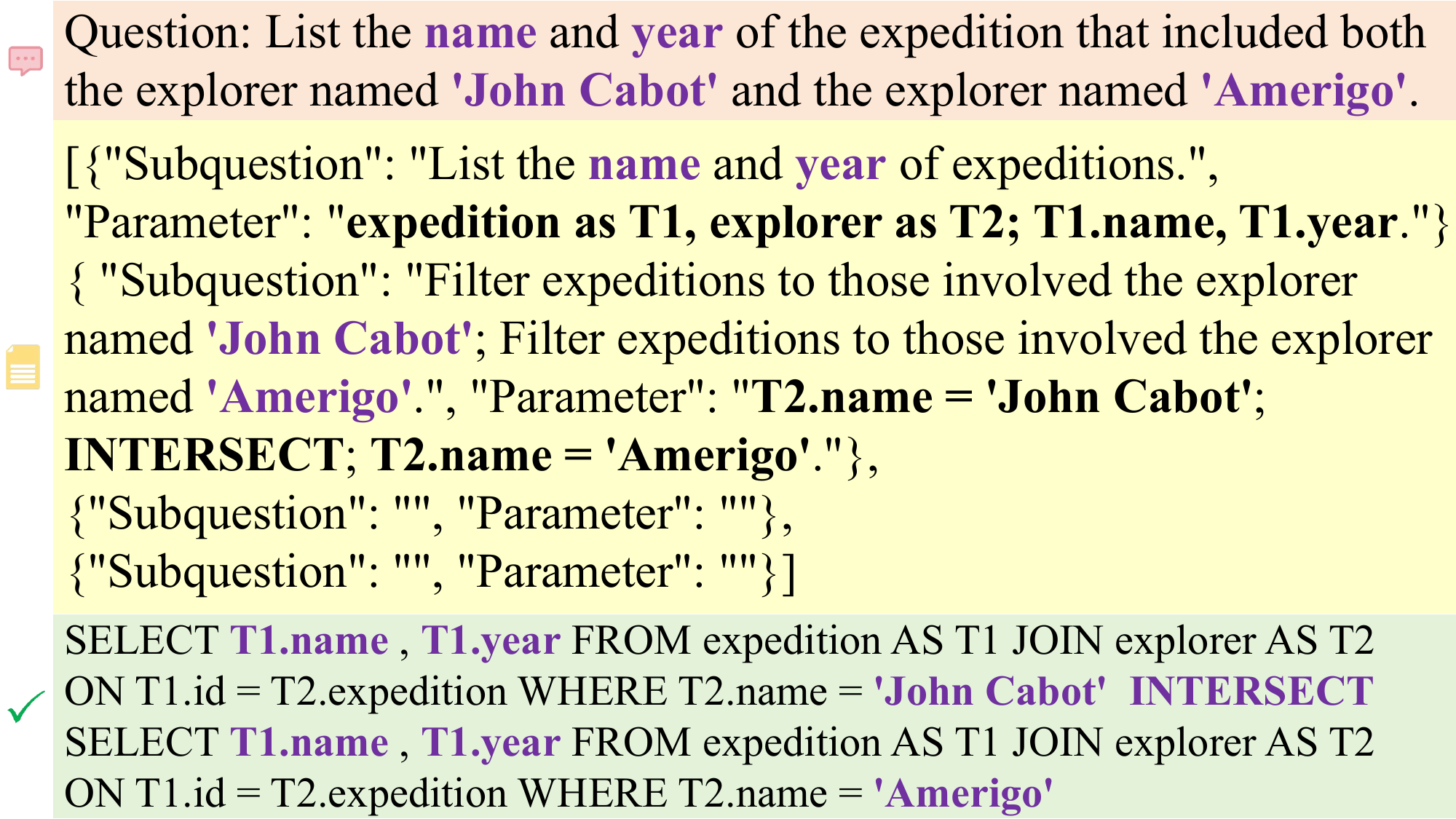}
        \caption{}
        \label{fig:case1}
    \end{subfigure}
    \hfill
    \begin{subfigure}[t]{0.495\columnwidth}
        \centering
        \includegraphics[width=\linewidth]{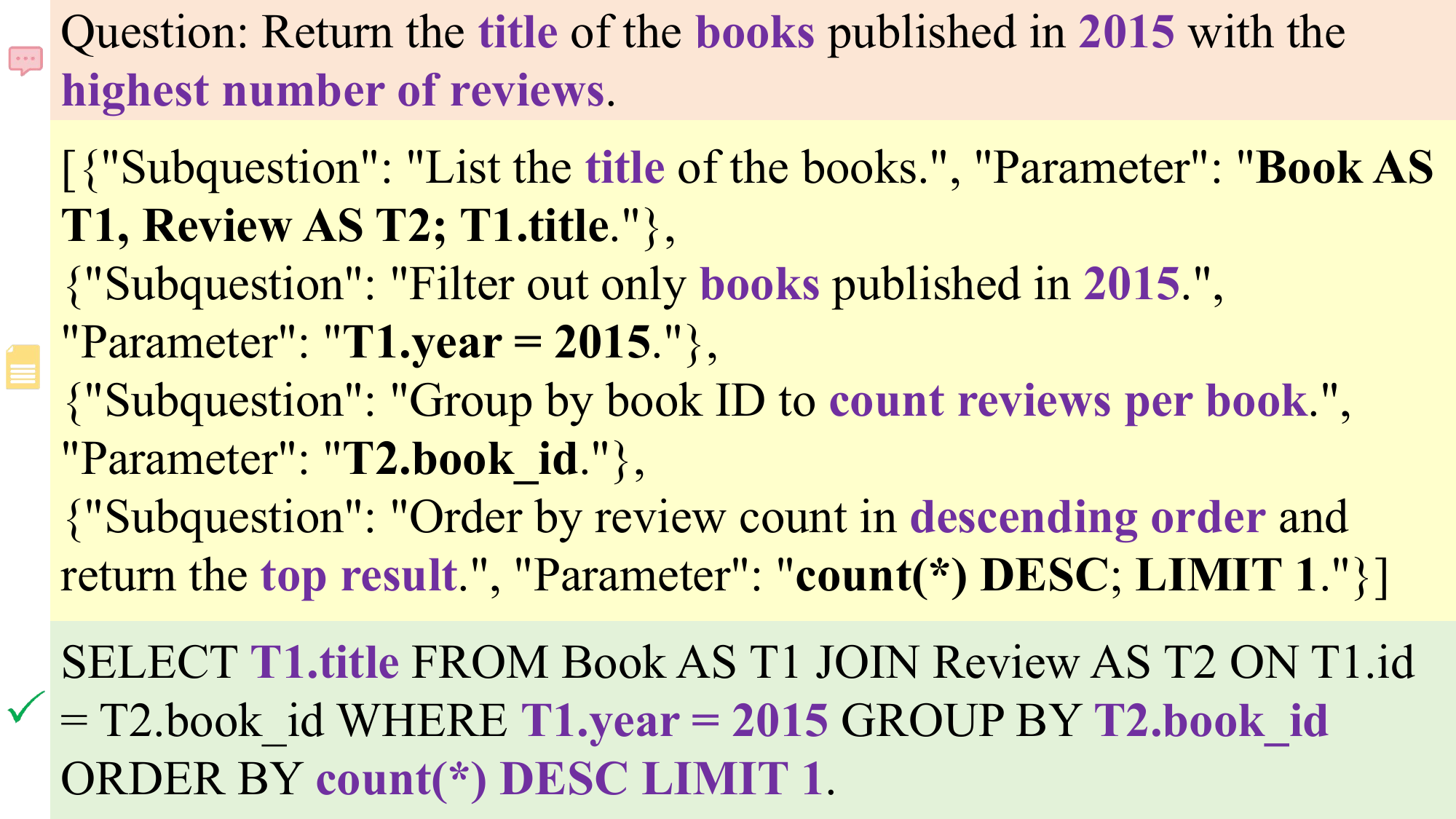}
        \caption{}
        \label{fig:case2}
    \end{subfigure}
    
    \caption{Case Study: (a) Decomposition of complex nested queries and set operations; (b) Modular parsing of multi-component SQL queries.}
    \label{fig:case}
    \vspace*{-8pt}
\end{figure}

We further illustrate the structure and workflow of SPOC-SQL by designing a case study shown in Fig. \ref{fig:case}, where SQL generation is decomposed into structured components with detailed sub-questions and corresponding parameters.
 Fig. \ref{fig:case} (a) illustrates Case 1, which involves complex nested subqueries and set operations,  
while Fig. \ref{fig:case} (b) presents Case 2, showing multi-component queries including joins, filtering, and grouping.  
In both cases, queries are split into sub-questions with corresponding SQL parameters, greatly reducing the complexity of generating accurate SQL and also helping users better understand the basis for the final SQL generation process.
\begin{figure}[t]
    \centering
    \vspace*{-12pt}
    \includegraphics[width=0.53\columnwidth]{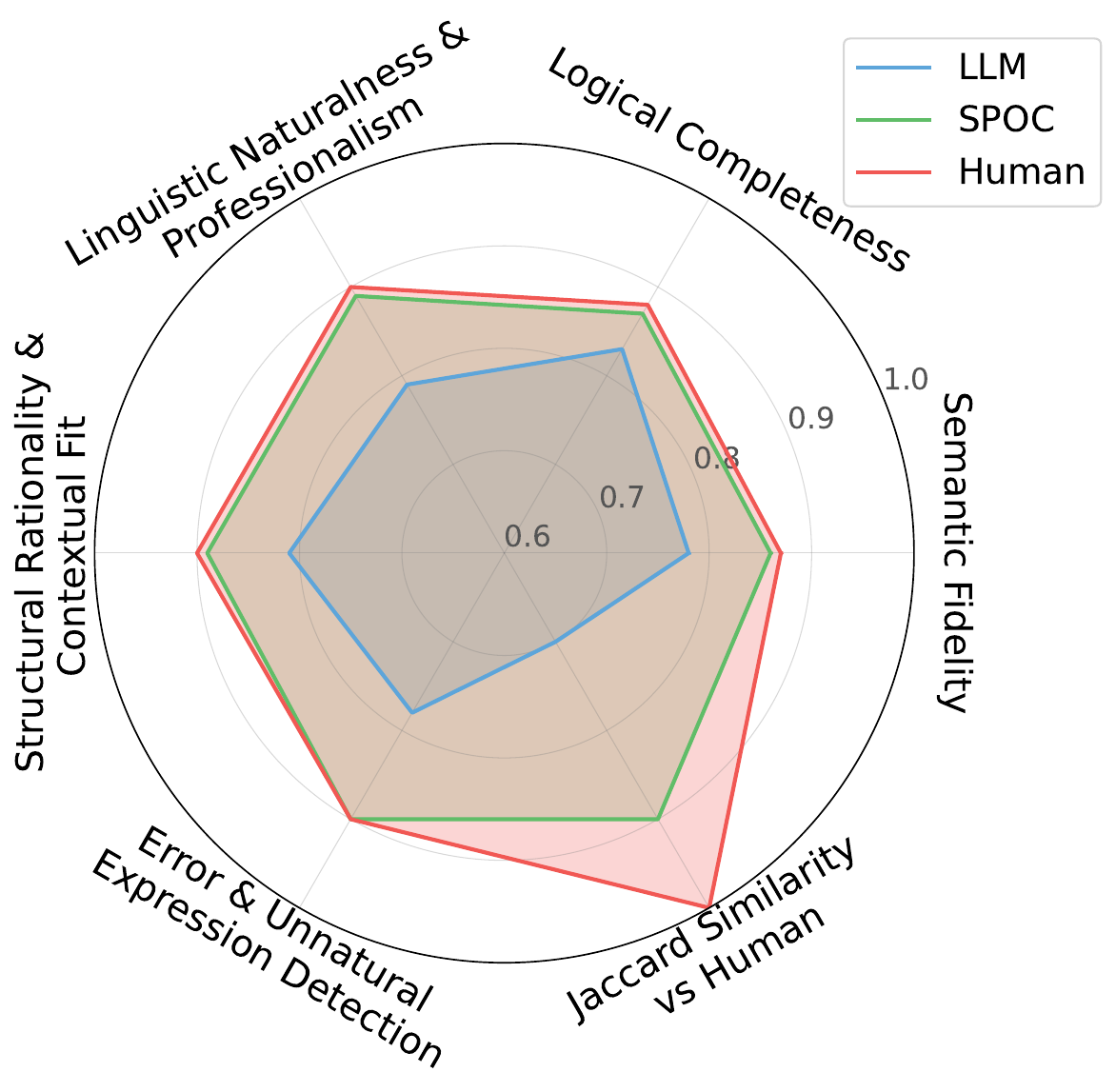}
    \caption{Comparison of SQL Info versions across subjective and objective metrics.}
    \label{fig:sqlinfo}
    \vspace*{-12pt}
\end{figure}

\vspace*{-15pt}
\subsection{User Study}
\vspace*{-6pt}
To validate whether the human intervention design in SPOC-SQL can effectively simulate real human intervention, we conducted a large-scale user study comparing three SQL Info versions: LLM, SPOC, and Human. Over 100 participants evaluated multiple subjective metrics on ten-point scales using samples from Spider-Dev and Spider-Realistic. As shown in Fig. \ref{fig:sqlinfo}, SPOC achieved scores close to Human and consistently outperformed LLM across all metrics. The correlation between subjective improvement and SQL accuracy gain was significant ($\rho = 0.6699$, $p < 0.01$), and the Jaccard Similarity vs Human metric further showed that SPOC outputs closely resemble human-authored content. These results demonstrate that the proposed intervention design can effectively simulate real human intervention.

\vspace*{-14pt}
\section{Conclusion}
\vspace*{-9pt}
In this paper, we propose SPOC-SQL, which effectively integrates multi-turn alignment procedures with preference optimization strategies for Text-to-SQL tasks. SPOC-SQL systematically decomposes single-turn queries into structured multi-turn supervision signals and incorporates PMDO to impose fine-grained preference constraints at key decision points across various SQL logical stages. Additionally, RDIM facilitates stage-wise decomposition and interactive verification, transforming SQL generation into a transparent, controllable, and collaborative process with correction capabilities at each generation phase. Extensive experimental results across multiple models and datasets demonstrate the effectiveness, robustness, and generalizability of the proposed method, underscoring its capacity to enhance both accuracy and interpretability in complex T2S tasks.

%
%
%

\bibliographystyle{splncs04}
\bibliography{Bibliography}

\end{document}